\documentclass[journal,review]{IEEEtran}

\usepackage{times}                     

\usepackage{booktabs}    
\usepackage{tabularx}    
\usepackage[caption=false,font=normalsize,labelfont=sf,textfont=sf]{subfig}

\usepackage[cal=cm]{mathalfa}

\usepackage{mathrsfs}
\usepackage{xcolor}
\usepackage{amsmath,amsfonts}
\usepackage{algorithmic}
\usepackage{algorithm}
\usepackage{array}
\usepackage{setspace}
\usepackage{textcomp}
\usepackage{stfloats}
\usepackage{url}
\usepackage{verbatim}
\usepackage{graphicx}
\usepackage[export]{adjustbox}
\usepackage{cite}
\usepackage{booktabs} 
\usepackage{cellspace}
\usepackage{amssymb}
\usepackage{amsfonts}
\usepackage{multirow}

\usepackage{tabu}                      
\usepackage{booktabs}                  
\usepackage{lipsum}                    
\usepackage{mwe}                       

\usepackage{mathptmx}                  

\usepackage{hyperref}
\hypersetup{
    colorlinks=true,      
    linkcolor=black,      
    citecolor=black,      
    urlcolor=blue,        
    allcolors=black       
}
\begin{document}

\title{Physically-guided Image Generation for Multi-Projection Mapping}

\author{Xingyun~Liu,   
 Yuqi~Li,
 Jinhui~Xiang,       Pinyan~Tang,
and~Chong~Wang
\IEEEcompsocitemizethanks{\IEEEcompsocthanksitem Xingyun Liu, Yuqi Li, Jinhui Xiang, Pinyan Tang, Chong Wang are with Ningbo University, Ningbo, 315211 P.R.China.  \protect\\
E-mail: 2411100088@nbu.edu.cn, liyuqi1@nbu.edu.cn, 2211100285@nbu.edu.cn,
tangpinyan@nbu.edu.cn, wangchong@nbu.edu.cn. \\
Corresponding Author: Yuqi Li}%
}



\maketitle

\begin{abstract}
Projection Mapping (PM) enables seamless superimposition of digital content onto real-world 3D objects, serving as a fundamental technique for immersive visualization, digital twins, and interactive art. Although text-to-image diffusion models have greatly facilitated customized content creation, directly integrating them into practical PM pipelines remains challenging due to the mismatch between idealized 2D generation and physical constraints. 

To bridge this gap, this paper formalizes two application-level generative paradigms: the cooperative paradigm (harmonizing generated semantics with physical attributes) and the adversarial paradigm (eliminating surface interference via radiometric compensation). Based on this, we propose ConPhyG, a unified controllable physically-guided generative multi-projection mapping framework that enables creators to interactively adjust physical constraints and flexibly switch generative paradigms. In cooperative mode, multi-dimensional physical priors (per-pixel gamut, depth, and edges) are injected into the diffusion process.  In adversarial mode, the framework releases the generative potential and applies bounded numerical optimization for multi-projector radiometric compensation. It allows users to dynamically switch constraints to balance artistic freedom with physical feasibility. Furthermore, we extend ConPhyG to 360-degree multi-view consistent PM using a sequential generation strategy. Quantitative and qualitative evaluations on a real-world four-projector setup demonstrate that ConPhyG significantly outperforms state-of-the-art methods in geometric alignment, gamut utilization, and semantic fidelity.  

\end{abstract}



\section{Introduction}
\IEEEPARstart{P}{rojection} Mapping (PM), also known as spatial augmented reality~\cite{bimber2005spatial}, has emerged as a powerful technology to augment the appearance of physical objects and environments by projecting virtual content onto real surfaces~\cite{shin2025recent, raskar2001shader}. It has found widespread applications in various fields, including industrial design, heritage exhibitions, and interactive entertainment~\cite{grundhofer2018recent, shin2025recent}.  Such augmentations are typically enabled by Projector-Camera systems (ProCams), integrated setups comprising one or more projectors and cameras~\cite{raskar1998office}. However, the content creation process for projection mapping remains constrained by high time and labor costs. The generation of high-quality projection content is highly dependent on manual customization by professional users, complex texture mapping, and tedious geometric alignment, which severely limits the scalability and widespread deployment of projection mapping systems in consumer scenarios.
 
Recent breakthroughs in generative artificial intelligence, particularly text-to-image diffusion models, have significantly lowered the barrier to content creation, making PM accessible to a broader range of users. Creators can now generate customized 2D textures using natural language prompts, bridging the gap between artistic intent and technical implementation. However, existing generative models~\cite{deng2025lapig,erel2023neural,deng2025gs}, relying solely on projection masks or RGB images of the projection surface, are fundamentally trained within an idealized 2D image space, largely unaccounted for the physical characteristics of real objects and complex environmental constraints inherent in real-world projector-camera systems (ProCams). When these idealized images are directly projected onto physical 3D objects with inconsistent geometry and reflectance, severe visual artifacts inevitably arise, including geometric misalignment, color clipping, and brightness compression. These mismatches are further amplified in multi-projector systems~\cite{majumder2025democratising}, where additional complexity is introduced to geometric alignment and color reproduction.

In this paper, we formalize a fundamental, previously unclarified dichotomy at the core of PM content generation: the Cooperative Paradigm versus the Adversarial Paradigm:
The \textit{Cooperative Paradigm} conceptually elevates the physical object from a passive display canvas to a positive component of the spatial visualization experience. It guides the generative pipeline to harmonize with the surface's inherent attributes, synthesizing digital graphics that appear to naturally emerge with the physical objects. In contrast,
the \textit{Adversarial Paradigm} treats the inherent physical properties of the object as undesirable visual interferences that must be eliminated. It completely decouples content generation from object attributes, synthesizing unconstrained creative imagery first in an idealized space, and subsequently relying on post-hoc radiometric compensation algorithms to remove visual artifacts and to closely approximate the target visual appearance.

A core insight of this work is that the applicability of these two paradigms critically depends on the degree of alignment between the creators' intentions (conveyed via text prompts) and the physical properties of the real object. This alignment can be decomposed into three independent dimensions: geometric alignment, color gamut alignment, and texture edge alignment. High alignment across all dimensions favors the cooperative paradigm, yielding naturally fused and view-robust projection results. By contrast, rigidly enforcing object constraints under low-alignment conditions severely distorts semantic intentions and degrades generative quality. In this case, the adversarial paradigm that generates content without constraints and applies the long-standing research classic radiometric compensation techniques becomes the only viable pathway to guarantee display fidelity.

To integrate the two complementary paradigms while retaining user artistic freedom, we propose a unified, \textbf{Con}trollable \textbf{Phy}sically-guided \textbf{G}enerative framework (ConPhyG) for multi-projection mapping. The framework provides a flexible interactive control interface, allowing creators to switch generation strategies based on the actual physical alignment status and creative intention. Under the cooperative mode, the framework explicitly injecting multi-dimensional object priors, such as per-pixel gamut, surface depth, and texture edges, into the early denoising stages of the diffusion process, ensuring that the generated semantics strictly adhere to the physical surface; Under the adversarial mode, the diffusion model performs unconstrained high-freedom generation, and a multi-projector radiometric compensation module based on constrained numerical optimization is invoked to compute the optimal input for each projector to eliminate radiometric interference of the surface. The key contributions of this work are summarized as follows: 

\begin{itemize}
\item We propose a unified, creator-controllable generative framework for multi-projection mapping, which supports flexible switching between cooperative generation paradigm and adversarial paradigm across diverse physical alignment scenarios. By embedding multi-dimensional physical priors into diffusion models and integrating them with a constrained numerical optimization-based radiometric compensation algorithm, our framework achieves a closed-loop optimization that balances semantic harmony with high physical fidelity.

\item We design a lightweight, novel In-Gamut Determination Network (IGDN) to efficiently classify whether target colors fall within the reproducible gamuts of multiple projectors in parallel, enabling gamut-aware content generation directly within the generative loop.

\item We conduct extensive quantitative and qualitative evaluations, including a comprehensive user study, on a real-world multi-projector ProCams setup to validate the framework's practical performance and enhanced creative freedom. Experimental results demonstrate robust compatibility with different textured objects, outperforming two state-of-the-art methods in setup adaptation time, display brightness, color fidelity, and operational flexibility.

\item  We present an initial exploration into extending our generative framework toward 360-degree multi-view consistent projection mapping. By introducing a sequential generation strategy, we offer an effective initial solution to inter-view inconsistencies and boundary artifacts, achieving multi-projector, multi-view, fully immersive spatial augmented reality.

\end{itemize}
\section{Related Work}
\label{sec:supplement_inst}

The research presented in this paper intersects geometric registration, radiometric compensation, and text-driven content synthesis within the ProCams domain.

\subsection{Geometric Registration}
Geometric registration is the process of establishing pixel-to-pixel correspondences between projector and camera coordinates to ensure that projected light reaches the desired position on the surface. Traditional registration methods rely on structured light (SL) patterns, such as Gray codes or phase-shifting fringes, which offer sub-pixel accuracy and robustness against environmental noise and inter-reflections~\cite{nayar2006fast}, particularly on non-planar~\cite{sajadi2011autocalibrating} or textured surfaces~\cite{moreno2012simple}.  To further enhance flexibility, researchers have explored imperceptible patterns, using additional invisible infrared (IR) light~\cite{hashimoto2017dynamic}, depth cameras~\cite{siegl2015real}, or synchronization triggers~\cite{cotting2004embedding} to enable geometric correction and projection simultaneously without visual interruption.  Alternative technical pathways include the use of physical markers (often invisible) attached to the projection surface~\cite{ibrahim2020dynamic}, which facilitate the augmentation of deformable or movable objects. By estimating the six degrees of freedom (6DOF) of the object, a view-independent projection mapping can be realized. Recent literature also leverages epipolar constraints~\cite{2021Radiometric} of ProCams to predict inter-pixel correspondence offsets in real-time. Furthermore, some approaches~\cite{yang2001automatic, li2012real} apply feature extraction and matching techniques to find sparse correspondences from natural images without extra devices, leveraging optical flow, such as GMA networks, to achieve real-time tracking during uninterrupted display processes.

\begin{figure*}[t] 
  \centering
  \includegraphics[width=\linewidth]{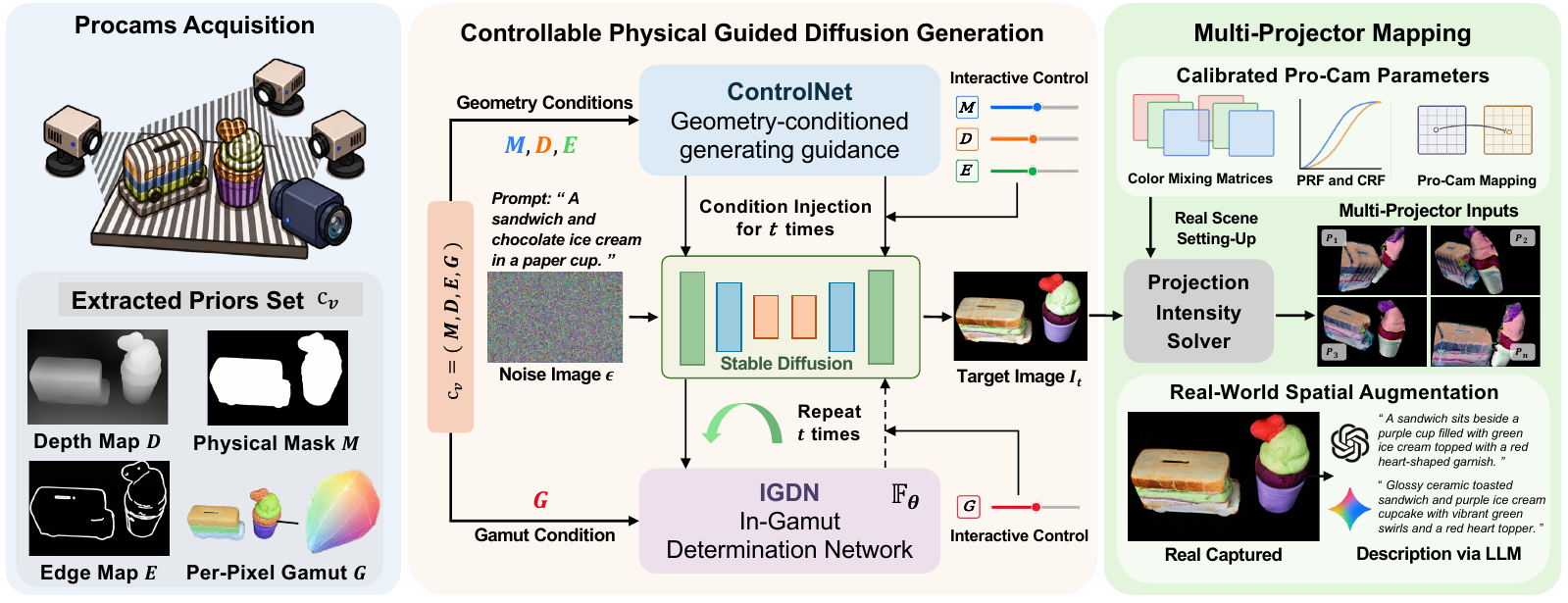}
  \caption{Overview of the proposed controllable physically-guided generative multi-projection mapping framework. The pipeline consists of three fundamental stages. \textbf{(Left) Priors Extraction:} The physical surface and projection mask ($\mathbf{M}$) are captured to extract geometric priors (Depth $\mathbf{D}$ and Edge maps $\mathbf{E}$). The per-pixel per-projector color-mixing($\mathbf{T}$) is captured to extract gamut priors $\mathbf{G}$. \textbf{(Middle) Controllable Physically-guided Generation:} During the $T$-step denoising loop, ControlNet enforces structural conditioning on the Stable Diffusion backbone. Simultaneously, the In-Gamut Determination Network(IGDN) module continuously computes physical gamut gradients to guide the intermediate latents ($X_t$), ensuring that the generated semantics strictly adhere to the gamut limitations of the real-world surface. \textbf{(Right) Multi-projector Image Calculation:} Bounded-Variable Least Square(BVLS) algorithm is used to calculate the pixel intensity for each projector. Finally, the calculated multiple projector images are projected onto the objects to create augmented effects.}
  \label{fig:pipeline}
\end{figure*}

\subsection{Radiometric Compensation}
Radiometric compensation aims to relieve the effects of arbitrary spatially varying surface reflectance and ambient light to ensure accurate color reproduction~\cite{bimber2005superimposing}. A fundamental method assumes a linear relationship in which the combined effect of the spectral power distribution (SPD) of projectors, the spectral reflectance of surfaces, and the spectral sensitivity function (SSF) of cameras is modeled by a $3 \times 3$ color mixing matrix~\cite{juang2007photometric}. This model is widely adopted for its computational efficiency. For systems requiring more complex modeling, such as DLP projectors with dependent channels~\cite{sajadi2010adict}, researchers have developed the compensation models involving dense sampling and high-dimensional interpolation functions~\cite{2015Robust}. To reduce the sampling time complexity, learning-based approaches leveraging manifold learning~\cite{2018Practical} and deep CNNs~\cite{huang2019end, 2020CompenNet,wang2024vicomp,wang2023compenhr,huang2021deprocams,huang2021end} are proposed to model these transformations implicitly as an alternative technical solution. Recent research has also explored physics-based factorization~\cite{li2023physics,deng2025gs}, achieving flexibility by decoupling static hardware factors (projector response functions) from dynamic scene properties like the spatially varying Bidirectional Reflectance Distribution Function (BRDF), enhancing interpretability and adaptability to changing surface conditions.

\subsection{Text-Generated Projection Mapping}
Recently, the integration of generative models has greatly enhanced the content creation potential of PM, with text-driven methods emerging as a user-friendly paradigm. Most existing approaches adopt a two-stage strategy: first, they leverage diffusion models to generate desired images based on the mask of the projection area; then, they create projector-compensated images to mitigate the effects of surface reflectance, utilizing either implicit Neural Radiance Fields (NeRF)~\cite{erel2023neural} or explicit Gaussian Splatting~\cite{deng2025gs} for 3D reflectance field representation. Casper DPM~\cite{erel2024casper} integrates 3D pose estimation with 2D deformation correction, focusing on articulated or moving human hands to achieve adaptive generation of text-guided content for dynamic surfaces. However, these methods generate desired images based solely on the projection area mask. While offering high generation flexibility, their radiometric compensation often suffers from saturation loss and brightness degradation due to inconsistencies between the generated desired images and the actual surface reflectance characteristics.
A style transfer-based approach~\cite{deng2025lapig} attempts to address this limitation by feeding RGB images of the projection surface into the generative model to guide desired image synthesis. Nevertheless, without explicitly modeling the projector's color gamut and the radiometric transformation between the projector and camera, these methods necessitate extensive clipping operations to ensure the generated projector images are numerically feasible, compromising the fidelity of the intended appearance. 

Furthermore, none of the aforementioned generation methods account for the depth or edge characteristics of projection surfaces, which is a key oversight that constrains their geometric and radiometric fidelity. Specifically, the lack of depth-aware generation constraints limits most approaches to single-view validity, failing to achieve true view-independent PM. Similarly, without explicit alignment between the edges of generated content and the surfaces, geometric mismatches are inevitable, which in turn increases the complexity of subsequent radiometric compensation. In addition, without 
reflectance-aware constraints, the models may generate highly saturated content on the surfaces with low reflectance in those spectral bands, leading to low responses after projection, and finally introduce out-of-gamut clipping or overall brightness compression.
Beyond these limitations, existing text-driven PM methods also lack the capability for multi-projection mapping, making them impractical for large-scale real-world applications that require seamless coverage of complex 3D surfaces~\cite{jones2014roomalive}.
These gaps highlight the need for a unified PM framework that bridges text-driven content generation with the physical realities of ProCam systems, while enabling practical and scalable multi-projection mapping.

\section{Methodology}

\subsection{Overview}

We consider a projection mapping (PM) system consisting of N projectors and one observation camera. For clarity, we first focus on this single-camera configuration throughout the core technical description; the extension to multi-camera systems for 360-degree view-independent projection mapping via a sequential generation strategy will be detailed in Section \ref{sec:360degree}. The $N$ projectors project virtual content onto the surface of a real object, with partially overlapping projection regions. The camera has a wider field of view (FOV) than each projector, such that it can observe the entire target object surface. In this work, we assume channel-independent LCD projectors, and the nonlinear radiometric response functions of both projectors and the camera are pre-calibrated to ensure linear projection and imaging behavior. The correspondence between projector pixels and camera pixels is also estimated in advance, thus establishing an accurate Procams.

Under such a static ProCams configuration, our goal is to generate the target camera-view image $\mathbf{I}_{target} \in \mathbb{R}^{H\times W\times3}$ that faithfully reflects the creative intention encoded by a natural language text prompt. As shown in Fig.\ref{fig:pipeline}, to ensure physical compatibility with the real ProCam system and the object surfaces, the pipeline incorporates three dimensions of physical constraints: \textit{depth}, \textit{texture edge}, and \textit{per-pixel gamut}, which will be formally defined in Section \ref{sec:constraint}. Depending on the degree of alignment between the prompt and the real physical attributes, the creators can flexibly toggle these constraints to dynamically switch between the cooperative and adversarial modes. Once the target camera image $\mathbf{I}_{target}$ is determined under the chosen configuration, we compute the optimal input intensity $\mathbf{P}_i$ for each pixel of the $i$-th projector, accounting for the light superposition in overlapping regions. The proposed physically guided generation schemes and multi-projector intensity optimization will be detailed in the following sections.

\begin{figure*}[t]
    \centering
    \includegraphics[width=0.8\linewidth]{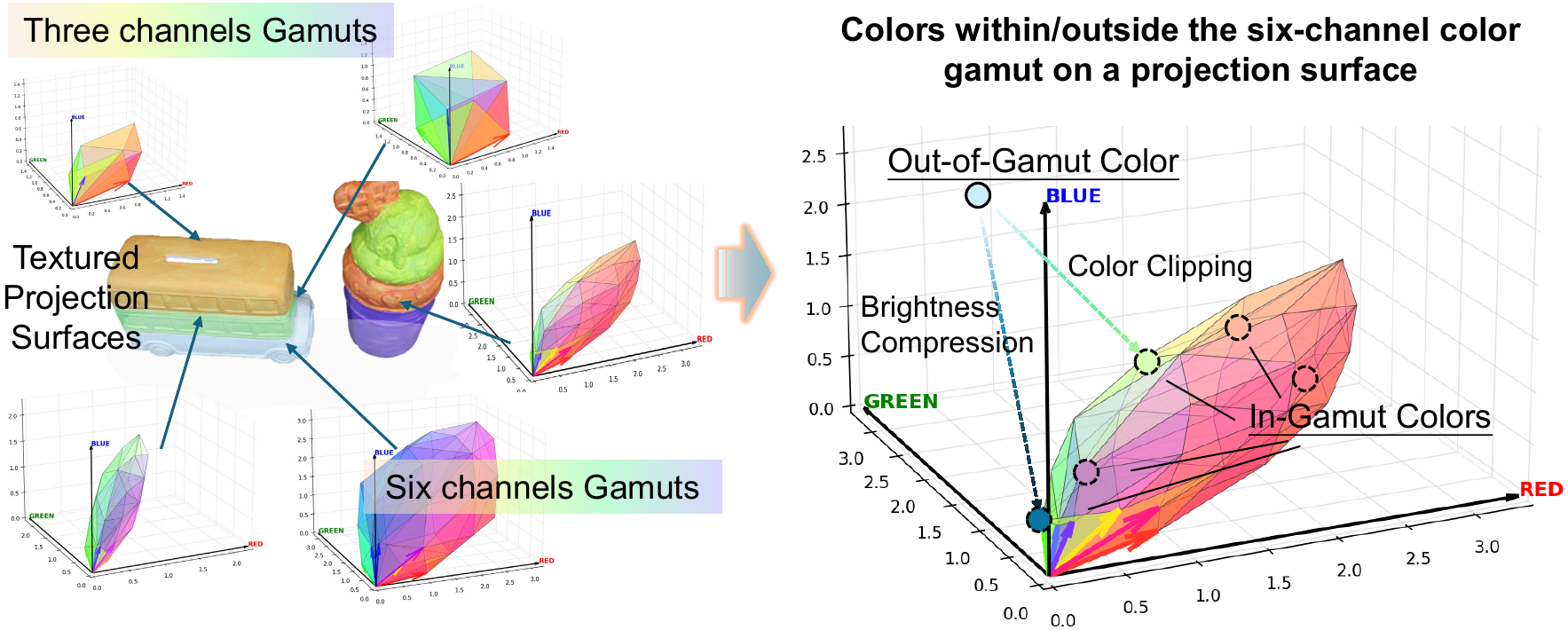} 
    \vspace{-1em}
    \caption{Visualization of per-pixel physical color gamuts. Left: Comparison of localized gamuts at various spatial positions on the physical object under a single-projector (3-channel) versus a dual-projector (6-channel) configuration. The per-pixel gamuts are modeled as zonotopes in the color space, geometrically spanned by the response vectors of each projector channel at that specific location, which are obtained from the ProCams calibration. Right: The brightness compression and color clipping effects caused by the out-of-gamut color.}
    \label{fig:gamut_comparison1}
\end{figure*}

\subsection{Projection Mapping Physical Constraint Definition}
\label{sec:constraint}
To embed physical feasibility constraints end-to-end into the text-driven generative pipeline, as shown in Fig.\ref{fig:pipeline}, we first define and pre-acquire four core physical priors of the ProCam system and target object, which serve as the hard constraints for our ConPhyG framework. All priors are defined from the camera’s viewpoint to ensure coordinate system consistency across the entire pipeline.
The physical priors include the surface depth map $\mathbf{D}$, geometric mask $\mathbf{M}$, texture edge  $\mathbf{E}$, and per-pixel gamut $\mathbf{G}$. 

\emph{Object Mask.} The mask $\mathbf{M}$ indicates the valid region of interest (ROI)  for PM. We extract $\mathbf{M}$ via background segmentation and binary thresholding on the neutral camera image of the static scene, to eliminate invalid background regions and to ensure that generated virtual content is strictly constrained within the projectable area of the target object.

\emph{Depth Map.} The depth map $\mathbf{D}$ describes the 3D geometric structure of the target surface. To acquire this geometric prior efficiently, we leverage monocular depth estimation via deep networks~\cite{ranftl2020towards,ranftl2021vision} rather than traditional hardware-based structured light scanning, as perceptual depth is more concerned. Specifically, we extract the dense depth map $\mathbf{D}$ directly from the camera-captured RGB image of the physical object using a pre-trained Dense Prediction Transformer (DPT) model. Constraining the generated content to be consistent with the perceptual depth map preserves semantic plausibility and mitigates perspective distortion when observed from typical viewing angles.

\emph{Texture Edges.} Texture edges $\mathbf{E}$ capture the inherent textural edges of the target object. We derive $\mathbf{E}$ from the captured image of the real objects using Canny edge detection~\cite{canny2009computational}, and aim to preserve the original texture details of objects during content generation rather than eliminate them. Perceptually, the human visual system exhibits acute sensitivity to edge alignments; enforcing edge consistency acts as a spatial anchor that binds the digital textures to the underlying geometry. 

\emph{Per-pixel Gamut.} 
The per-pixel gamut $\mathbf{G}$ models the reproducible color range of each camera pixel under the multi-projector system configuration. Without loss of generality, consider a pixel $\mathbf{I}(u) \in \mathbb{R}^3$ in the captured image by the camera, where $u$ denotes the pixel coordinate in the plane of the camera image. The response of this pixel is the combined result of the color intensity of overlapping projector pixels $\{\mathbf{P}_i(u_i') \in \mathbb{R}^{3\times 1} |i=1,\dots, N\}$, the spectral radiance basis $\mathbf{L}(u_i')$ of the $i$-th projector, the surface spectral reflectance $\mathbf{R}(u)$ at the 3D position corresponding to camera pixel  $u$, and the intrinsic spectral sensitivity function(SSF) $\mathbf{S}$ of the camera. Here $\{u_i'|i=1,\dots, N\}$ are the corresponding pixel coordinates in the image plane of the $i$-th projector $\mathbf{P}_i$, established via geometric calibration of the camera-projector system. Previous works~\cite{grossberg2004making, 2018Practical} have validated the linear color-mixing model for such systems, formulated as:
\begin{equation}
\mathbf{I}(u) = \sum_{i=1}^{N} \mathbf{T}_i(u) \mathbf{P}_i(u_i'), 
\end{equation}
where $\mathbf{T}_i(u)$ is the per-pixel $3 \times 3$ color mixing matrix of the $i$-th projector at camera coordinate $u$. $\mathbf{T}_i$ is the result of multiplication of the radiance basis $\mathbf{L}(u_i')$ of the $i$-th projector, reflectance $\mathbf{R}(u)$ in pixel $u$ and camera SSF $\mathbf{S}$. Note that if the $i$-th projector has no contribution to the camera pixel $u$, the value $\mathbf{T}_i = 0$.
Since the valid intensity range of each projector pixel is limited to $\mathbf{0} \leq \mathbf{P}_i(u_i') \leq \mathbf{1}$, the reproducible color set for the camera pixels can be defined as: 
\begin{equation}
\mathbf{G}(u) = \{ \sum_{i=1}^N \mathbf{T}_{i}(u) \mathbf{P}_i(u_i')| 0 \leq \mathbf{P}_i(u_i') \leq 1 \}. 
\end{equation}
Therefore, the per-pixel gamut is a zonotope-like convex polytope(as shown in Fig.\ref{fig:gamut_comparison1}), constructed in a per-channel, per-projector, and per-pixel manner. We acquire the color-mixing matrix map $\mathbf{T}$ by projecting single uniform color (R, G, B) images onto the target surface. With pre-calibrated linear response functions of the projectors and camera, the per-pixel gamut can be established.

All the above physical priors are pre-acquired before the generative process, and provided as structured physical guidance for our desired image generation model.

\subsection{Physically Guided Generation}

Our physically guided text-driven generation framework for the desired camera images is built upon a diffusion model SDXL~\cite{podell2024sdxl},  where a Markovian forward process \(q(x_t|x_0)\) gradually injects Gaussian noise into clean input data \(x_0\) to train the model to recover clean signals from noisy inputs. This forward process is defined as:
\begin{equation}
x_t = \sqrt{\bar{\alpha}_t} x_0 + \sqrt{1 - \bar{\alpha}_t} \epsilon, \quad \epsilon \sim \mathcal{N}(\mathbf{0}, I),
\label{eq:diffusion}
\end{equation}
where \(\bar{\alpha}_t = \prod_{s=0}^t \alpha_s\), \(\alpha_t = 1 - \beta_t\) is a timestep-dependent function determined by the denoising sampler (e.g., DDPM), and \(\epsilon\) is a noise map sampled from a standard normal distribution. The core training objective of the diffusion model is to minimize the mean squared error between the predicted noise \(\epsilon_\theta\) (output by a U-Net with parameters \(\theta\)) and the true noise \(\epsilon\).
For controllable generation tailored to projection mapping, we encode the above physical constraints, as structured image conditions \(c_v=(\mathbf{M}, \mathbf{D}, \mathbf{E}, \mathbf{G})\), alongside natural language text prompts \(c_t\). These physical priors are embedded via ControlNet~\cite{zhang2023adding}, which guides the U-Net denoising process to ensure generated content adheres to real-world geometric and radiometric constraints. The diffusion training loss is re-formulated to incorporate these conditions:
\begin{equation}
\mathcal{L}_{\text{train}} = \mathbb{E}_{x_0, t, c_t, c_v, \epsilon \sim \mathcal{N}(0, 1)} \left[ \left\| \epsilon_\theta(x_t, t, c_t, c_v) - \epsilon \right\|_2^2 \right].
\label{eq:condition}
\end{equation}
During inference, starting from random noise \(x_T \sim \mathcal{N}(\mathbf{0}, I)\), the model reconstructs the final denoised image \(x_0\) via a step-by-step denoising process, where each step refines the output while respecting the embedded physical constraints.

To further enforce alignment between generated content and physical constraints, we integrate a consistency feedback mechanism inspired by ControlNet++~\cite{li2024controlnet++}. A discriminative reward model \(\mathbb{D}\) extracts corresponding physical conditions (e.g., masks, edge maps, depth estimates, color gamut projections) from the generated images \(x_0'\), and we compute a reward loss to quantify the consistency between these extracted conditions \(\hat{c}_v\) and the input physical constraints \(c_v\). This reward consistency loss is defined as:
\begin{equation}
\mathcal{L}_{\text{reward}} = \mathcal{L}\left(c_v, \hat{c}_v\right) = \mathcal{L}\left(c_v, \mathbb{D}\left(x_0'\right)\right) = \mathcal{L}\left(c_v, \mathbb{D}\left[ \mathbb{G}^T(c_t, c_v, x_T, t) \right]\right),
\label{eq:rewardloss}
\end{equation}
where \(\mathbb{G}^T(c_t, c_v, x_T, t)\) denotes the full \(T\)-step denoising process that generates \(x_0'\) from random noise \(x_T\), and \(\mathcal{L}\) is a task-specific metric. We define the corresponding task-specific loss terms under the unified framework of Eq.(\ref{eq:rewardloss}), with detailed descriptions as follows:

\emph{Mask Loss.} We extract the corresponding mask map $\hat{c}_v^{m} = \mathbb{D}(x_0')$  from the generated image $x_0'$. For the reward loss function term, we adopt per-pixel binary cross-entropy (BCE) loss for binary masks as the task-specific metric. The loss is formulated as:
\begin{equation}
\mathcal{L}_{reward}^{m} = \mathcal{L}_{bce}(\mathbf{M}, \hat{c}_v^{m}).
\end{equation}
This term ensures that the spatial region of the real object is consistent with that of the generated image.

\emph{Depth Loss.} We extract depth map $\hat{c}_v^{d} = \mathbb{D}_{d}(x_0')$  from the generated image $x_0'$. For the depth loss term,  we adopt mean squared error (MSE) regression loss as the task-specific metric. The loss is formulated as:
\begin{equation}
\mathcal{L}_{reward}^{d} = \mathcal{L}_{mse}(\mathbf{D}, \hat{c}_v^{d}).
\end{equation}
This term ensures the 3D spatial structure and geometric realism of the generated content comply with the input physical depth constraints.

\emph{Edge Loss.} The edge consistency loss measures the structural alignment between the input binary edge map constraint $\mathbf{E} \in \{0,1\}^{H\times W}$ 
 (where 1 denotes valid edge pixels and 0 denotes non-edge pixels) and the edge probability map $\hat{c}_v^e = \mathbb{D}_e(x_0')\in \{0,1\}^{H\times W}$ extracted from the generated image $x_0'$.  To alleviate the pressure of the post radiometric compensation operation, we explicitly force that the structural edge of the objects must be preserved in the generated images. To focus on the coincidence of edge structures, we design the loss based on the intersection of the two edge maps, formulated as:
\begin{equation}
\mathcal{L}_{reward}^e = 1-\frac{|\mathbf{E} \cap \hat{c}_v^e|}{|\mathbf{E}|}.
\end{equation}
Minimizing this loss term forces the structural edge of the real objects to appear in the generated images, while the reverse does not necessarily hold. This design yields more natural results for the final projection mapping and effectively reduces the difficulty of the radiometric compensation step.

\emph{Gamut Loss.} The gamut loss enforces that the color of each pixel in the generated camera image lies within the pre-calibrated per-pixel gamut. Conventionally, determining whether a pixel color falls inside or outside its corresponding gamut requires solving a constrained numerical optimization problem \cite{li2015content, morovic2001fundamentals}, which introduces non-negligible computational complexity and is incompatible with end-to-end differentiable optimization in the generation loop. To accelerate this process, we design a neural network $\mathbb{F}$ to determine colors inside or outside the gamut in a differentiable manner. Full architectural and implementation details of $\mathbb{F}$ are presented in section \ref{sec:gamut_net}. The output of this network is directly adopted as the gamut loss term in our generation framework, formulated as:
\begin{equation}\mathcal{L}_{\text{reward}}^g = \mathbb{F}\left( \mathbf{G}, \hat{c}_v \right),\label{eq:gamutloss}\end{equation}
This term penalizes out-of-gamut pixel colors, and ensures that all colors in the generated image are physically reproducible under the real object surfaces and projector-camera (PM) system configuration, avoiding color clipping or brightness compression (as shown in Fig.\ref{fig:gamut_comparison1}) in the radiometric compensation step.

By combining this reward consistency loss with the standard diffusion training loss into a unified objective, we balance two critical goals:
\begin{equation}
\mathcal{L}_{\text{total}} = \mathcal{L}_{\text{train}} + \lambda_m  \mathcal{L}^m_{\text{reward}}+ \lambda_d  \mathcal{L}^d_{\text{reward}}+ \lambda_e  \mathcal{L}^e_{\text{reward}}+ \lambda_g  \mathcal{L}^g_{\text{reward}},
\label{eq:totalloss}
\end{equation}
where \(\lambda\) are hyperparameters that weight the contribution of each reward loss. This formulation ensures that the model preserves its strong text-to-image semantic generation capabilities while being explicitly guided to produce content that is geometrically consistent, color-accurate, and directly deployable in multi-projector systems, closing the gap between idealized generative outputs and the practical constraints of real-world ProCam setups. In the training stage, rather than conducting extensive retraining from scratch, we perform targeted fine-tuning of ControlNet to implicitly align multi-dimensional physical priors with the diffusion model's latent space. During the inference stage, instead of relying solely on feed-forward generation, we introduce an interactive test-time guidance strategy to iteratively refine intermediate latents at runtime. Specifically, at each denoising step $t$, the gradients of the physical reward losses are backpropagated to dynamically update the noisy latent $\mathbf{x}_t$. Governed by user-specified constraints, this step-by-step refinement explicitly steers the generative trajectory towards a physically feasible solution space.

\subsection{In-Gamut Determination Network}
\label{sec:gamut_net}

In this section, we introduce our In-Gamut Determination Network (IGDN)  $\mathbb{F}_\theta$, a differentiable module designed to quantify whether a desired RGB color $\mathbf{c} \in \mathbb{R}^3$ lies within or outside the zonotope-like per-pixel gamut $\mathbf{G}$. Leveraging the fact that the per-pixel gamut is explicitly constructed via a linear color mixing model, the network enables efficient in-gamut determination, while supporting end-to-end backpropagation of the physical feasibility gradient into the latent space of our diffusion model.
The boundary of $\mathbf{G}$ is fully defined by the color-mixing tensor $\mathbf{T} \in \mathbb{R}^{N\times 3 \times 3}$. Therefore, we use 
$\mathbf{T}$
as one of the network inputs, and the 
three-dimensional desired RGB color vector $\mathbf{c}$ as the second input. The network outputs a single non-negative scalar value, which quantifies the minimal Euclidean distance from the desired color to the feasible gamut set. An output value of 0 indicates that the target color is predicted to lie within the valid gamut defined by $\mathbf{T}$; a positive output value corresponds to the minimal color deviation required to approximate the target color $\mathbf{c}$ 
 under the physical constraints.
 
The IGDN is implemented as a Multi-Layer Perceptron (MLP) $\mathbb{F}_\theta$ with six layers and ReLU activation functions, where $\theta$ denotes the trainable parameters. The last ReLU layer is applied to strictly enforce the non-negativity of the network output, consistent with the physical definition of color reproduction error.
Training the IGDN requires a large-scale, balanced dataset covering both in-gamut and out-of-gamut samples. To construct the training dataset, we adopt a uniform sampling strategy to sample colors in the full normalized intensity space $[0,1]^{3}$ and uniformly sample the $N$ color-mixing matrix around the $3 \times 3$ identity matrix for diagonal domination. The training set consists of 30,000 samples of $(\mathbf{c}^j, \mathbf{T}^j; y^j)$, where $y^j$ denotes the $j$-th color reproduction error of using $\mathbf{T}^j$ to reproduce $\mathbf{c}^j$. The value of the label $y^i$ is obtained by applying Gurobi to solve the constrained optimization problem:
\begin{equation}
y^j = \min ||\mathbf{c}^j - \sum_{i=1}^N \mathbf{T}^j_i \mathbf{\alpha}_i||_2,\\
\quad \text{s.t.} \; \mathbf{\alpha}_i \in \mathbb{R}^{3\times 1}, \mathbf{0} \leq \mathbf{\alpha}_i \leq  \mathbf{1}.
\end{equation}
Once trained, the IGDN is frozen and integrated into the diffusion generation pipeline as the gamut loss function. The generative process is then guided by backpropagating the gradient of the gamut loss, and the gradient effectively pushes the pixel colors of the generated image toward the valid gamut, ensuring the final generated image is physically reproducible under the real camera-projector system configuration while preserving the semantic content of the generation target.

\subsection{Multi-projector Intensity Calculation}
\label{sec:intensity_calc}

Once a physically feasible target camera image $\mathbf{I}_{target}$ is generated, the final step is to accurately solve for the projector input intensities $\mathbf{P}_i$. While the IGDN ensures the existence of a valid solution, it does not output the intensities of each channel. To achieve pixel-level radiometric compensation and seamless blending across overlapping regions, we formulate this as a constrained linear least-squares optimization problem.

For each pixel $u$, we seek the optimal intensity vector $\mathbf{P}(u') = [\mathbf{P}_1(u_1'), ..., \mathbf{P}_N(u_N')]^T$ that minimizes the reconstruction error between the simulated projection and the target image, subject to the hardware's dynamic range limits:
\begin{equation}
    \min_{\mathbf{P}} \left\| \left( \sum_{i=1}^{N} w_i(u_i') \mathbf{T}_i(u) \mathbf{P}_i(u_i') \right) - \mathbf{I}_{target}^{linear}(u) \right\|^2_2
    \quad \text{s.t.} \quad 0 \le \mathbf{P}_i(u_i') \le 1
\end{equation}
Here, $\mathbf{I}_{target}^{linear}$ is the target image transformed from sRGB to the linear radiometric space via Gamma decoding ($\gamma=2.2$) to match the physical linearity of light superposition. The term $w_i(x)$ represents a normalized soft-blending weight derived from the distance transform of the projection masks, ensuring smooth transitions in overlapping zones.

This optimization is solved efficiently using the Bounded-Variable Least Squares (BVLS) algorithm. Unlike neural inference, this mathematical solver guarantees that the calculated inputs strictly adhere to physical bounds (avoiding clipping artifacts) and optimally distributes brightness among overlapping projectors~\cite{stone1988color}. Finally, the solved linear intensities $P_i$ are mapped to the projector's native input space (0-255) using pre-calibrated inverse radiometric response functions (LUTs).

\subsection{Towards 360-Degree View-Independent Projection}
\label{sec:360degree}
The previous sections detailed our method for single-view projection mapping. As truly immersive spatial augmentation inherently requires comprehensive 360-degree coverage of physical objects, and generative solutions for consistent multi-view projection mapping remain largely underexplored, we conduct a preliminary extension of our proposed framework. Specifically, we experiment with a sequential generation strategy as an initial approach to alleviating inter-view inconsistencies.

The sequential generation strategy can be applied to multiple viewpoints and operates in an incremental manner. This strategy relies on the assumption that there are overlapping regions between adjacent viewpoints, and these overlapping regions provide the necessary common reference for cross-view alignment.
We first optimize the target and projected images for the first $k$ views. Once the virtual content has been successfully mapped to the corresponding surface regions, we proceed to process the $(k+1)$-th adjacent viewpoint, which shares overlapping regions with the first $k$ viewpoints. At this stage, all projectors maintain the projection content that has already been optimized for all previous viewpoints. We extract a new projection mask from the $(k+1)$-th view, together with the partial projection of the object captured by that camera, and feed them as another control condition into our ControlNet to generate the expected image for the $(k+1)$-th view. This sequential conditioning mechanism leverages the overlapping regions to facilitate structural and photometric alignment between the content generated for the new view and the existing projections on the physical object.


\begin{figure}[htp]
  \centering
  \includegraphics[width=\linewidth]{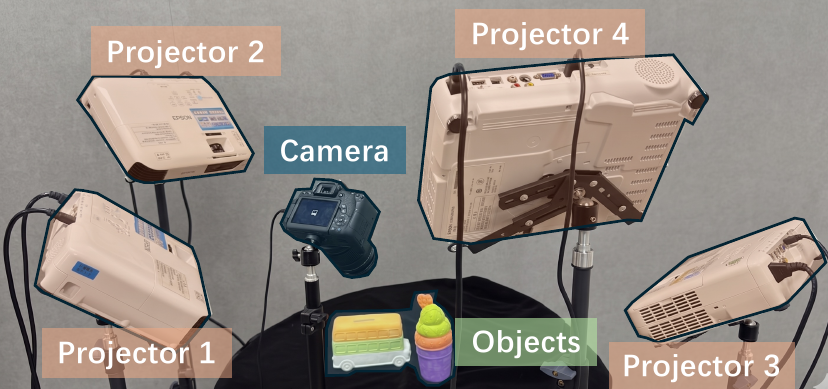}
  \caption{Our experimental setup consists of four projectors and one camera. The projection and capturing areas fully cover the object.}
  \label{fig:setup}
\end{figure}

  
  

\section{Experiments}

\subsection{Experiment Setup}
Our experimental system comprises four Epson CB-X31 3LCD projectors and a Canon EOS 750D DSLR camera equipped with an 18-135mm zoom lens, as shown in Fig.\ref{fig:setup}. 
The projectors were configured to operate at a resolution of $1024 \times 768$ pixels, while the camera captured images at a resolution of $6000 \times 4000$ pixels. 
The target objects used in the experiments consisted of multiple plaster and plastic models, featuring a wide variety of geometric structures, which were partially painted with pigment to introduce surface color variations. 
All experiments were run on a central workstation equipped with an Intel Core i3-13100F CPU, 64 GB of RAM, and an NVIDIA GeForce RTX 4090 GPU.

\subsection{Calibration}
We performed a one-time end-to-end system calibration to establish accurate geometric and radiometric correspondences between all projectors and the camera.
For geometric calibration, we projected standard binary Gray code structured light patterns onto the target object to obtain sub-pixel-level pixel-to-pixel mapping between each projector and the camera.
For radiometric response calibration, we projected a sequence of uniform grayscale images with intensities ranging from 0 to 255 in steps of 10 onto a neutral white reference surface. We then captured these images with the camera and fitted a third-order polynomial function to calibrate the nonlinear radiometric response of both the projectors and the camera, ensuring a linear relationship between input pixel values and output radiance.
For color mixing matrix estimation, we projected pure red, green, and blue uniform images sequentially. By analyzing the camera-captured responses of these primary colors, we computed a per-pixel $3\times 3$ color mixing matrix for each projector-camera pair, which accounts for spatial variations in color response across the projector's field of view.

\subsection{Workflow}

\begin{figure}[h]
  \centering
  \includegraphics[width=\linewidth]{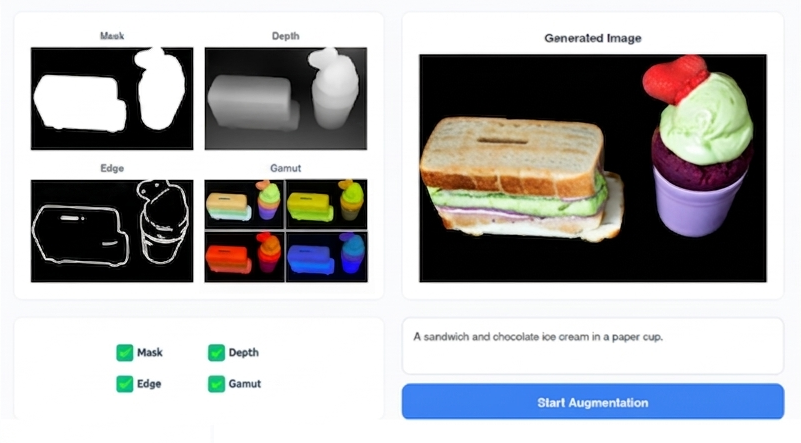} 
  \caption{Our user interface that includes a prompt input box and four checkboxes for enabling constraints.}
  \label{fig:ui_for_users}
\end{figure}

To generate content for multi-projection mapping applications, we first used the calibrated Procams to capture the target physical object and extract four physical priors: mask, depth, edge, and per-pixel gamut. All extracted priors are presented on the interactive interface for user preview. As shown in Fig.\ref{fig:ui_for_users}, creators then input a text prompt to specify their creative intentions, and manually enable or disable each physical prior based on the degree of alignment between their desired content and the object’s inherent properties. Activating a constraint engages the cooperative paradigm along the corresponding dimension for seamless physical-digital fusion, while disabling all priors shifts the pipeline to the adversarial paradigm for unconstrained creative freedom.
Upon triggering the 'Start Augmentation' button, the pipeline injects the selected priors into the diffusion model to generate the target projection content. Users could iteratively tune the constraint combinations to refine the output. Once the generated result is confirmed, the system computes per-projector input images for final projection. In overlapping regions of adjacent projectors, a standard linear alpha-blending strategy is adopted to ensure smooth luminance at the projection overlapping regions.

\begin{figure*}[htp]
  \centering
  \includegraphics[width=1.01\linewidth]{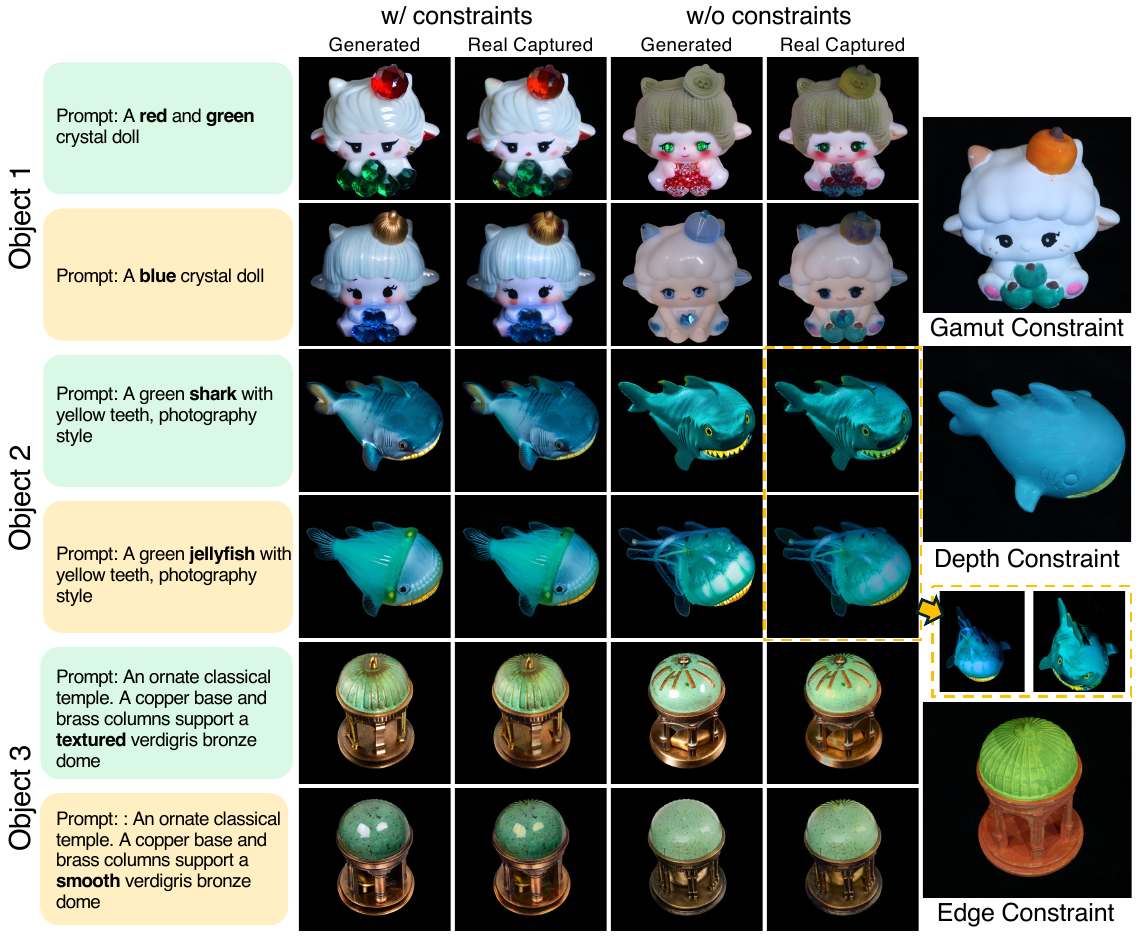}
\caption{Qualitative evaluation of individual physical constraints under varied object-prompt combinations. Both generated and real captured projection results are presented. Green and yellow boxes denote prompts that are consistent and inconsistent with the target object, respectively. Individual constraints are isolated to evaluate their distinct impacts (with other constraints active): gamut constraint for Object 1, depth constraint for Object 2, and structural edge constraint for Object 3. Note that for the mismatched combination (jellyfish prompt with shark object in Object 2), disabling the depth constraint improves semantic fidelity but severely degrades appearance quality from novel viewpoints.}
  \label{fig:3 constraints}
\end{figure*}%

\begin{figure*}[htp]
  \centering
  \includegraphics[width=\linewidth]{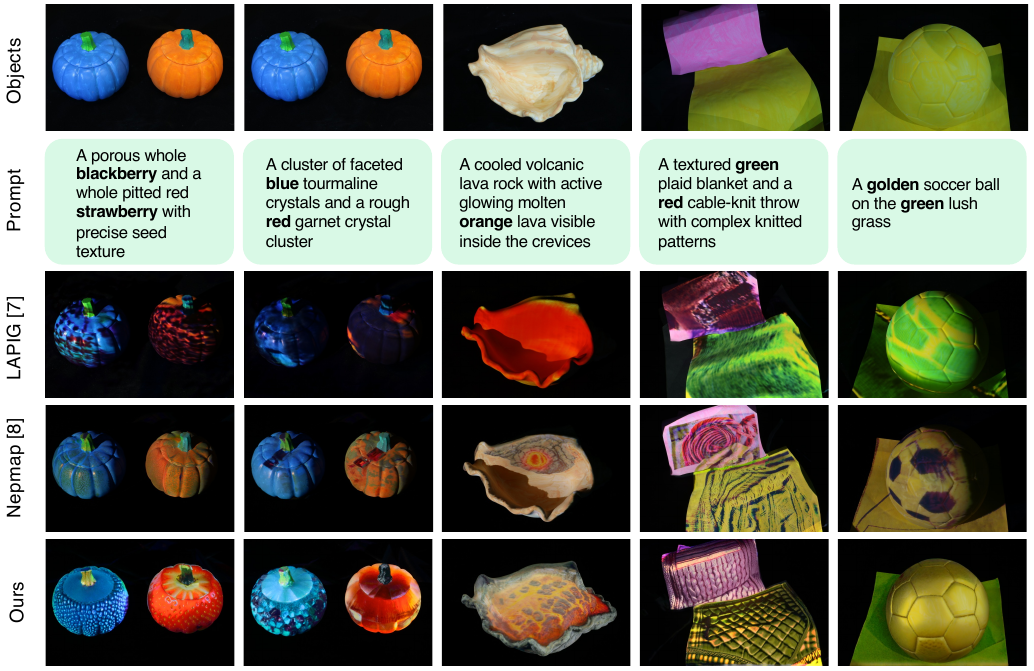}
  \vspace{-1em}
  \caption{Visual comparison of projection results between LAPIG~\cite{deng2025lapig}, NepMap~\cite{erel2023neural}, and our method on four representative objects: pumpkins, conch, curved paper, and soccer ball. Our method achieves superior overall visual performance, including: (1) accurate semantic consistency; (2) precise geometric alignment (evident in the soccer ball seams); (3) excellent virtual-real surface depth consistency (notable in the conch and curved paper cases); (4) higher brightness and richer color fidelity (clearly demonstrated in the pumpkin cases).}
  \label{fig:rcomparison esult}
\end{figure*}

By toggling the geometric and color gamut constraints, our framework is fully capable of handling diverse scenarios characterized by either consistency or discrepancy between the input text prompts and the real objects. In Fig.~\ref{fig:3 constraints}, we present the generated results and final multi-projection mapping effects under various prompt-object pairs across different constraint combinations. Visual results demonstrate that when the reflectance, semantic, and texture attributes of the target object exhibit high alignment with the input prompt, the explicit injection of color gamut, depth, and structural priors achieves superior visual fidelity. However, in low-alignment cases where the physical characteristic contradicts the prompt semantics, enforcing these constraints versus relaxing them presents a clear trade-off. Our results indicate that rigidly maintaining the cooperative paradigm under such mismatches often introduces unexpected and undesirable semantic artifacts that compromise the creator's original intention, as shown in the 'jellyfish' case on the 'shark' object with depth constraint in Fig.\ref{fig:3 constraints}. Therefore, transitioning to the adversarial paradigm is highly recommended in these scenarios to facilitate unconstrained generation, but this transition sacrifices other properties such as brightness and view-independence. In such cases, creators need to choose the optimal generated images for the next step (physical projection mapping).

\subsection{Comparison with State-of-the-Art Methods}

\subsubsection{Objective Comparison}
We conducted our comparative analysis directly on the actual captured images of the physical projections mapping. Unlike idealized numerical simulations, physical captures more accurately reflect the generation method's capabilities for handling real-world geometric misalignments, spatially varying surface spectral reflectance, projector gamut limitations, and complex multi-projector blending. 

\begin{table}[tb] 
    \centering
    \caption{The comparison of projection quality across different methods. ($\uparrow$ indicates higher is better, $\downarrow$ indicates lower is better). Best results are highlighted in \textbf{bold}.}
    \label{tab:quantitative_results}
    \resizebox{\columnwidth}{!}{%
    \begin{tabular}{lccccc} 
        \toprule
        \textbf{Method} & \textbf{CLIP} $\uparrow$ & \!\textbf{BLIP ITM}\! $\uparrow$\! & \textbf{PSNR} $\uparrow$ & \textbf{$\Delta E$} $\downarrow$ &\;\; \textbf{Clipping Error} $\downarrow$\;\;\\
        \midrule
        LAPIG~\cite{deng2025lapig} & 23.83 & 16.5\% & 10.80 & 31.78 & 0.174\\
        Nepmap~\cite{erel2023neural} & 23.01 & 17.2\% & 12.31 & 14.73 & 0.255\\
        Ours & \textbf{26.37} & \textbf{78.9\%} & \textbf{22.38} & \textbf{4.69} & \textbf{0.039}\\
        \bottomrule
    \end{tabular}%
    }
\end{table}

\textbf{Qualitative Comparison.} Fig.\ref{fig:rcomparison esult} illustrates the visual comparisons between our method and existing text-driven projection mapping baselines. When projecting onto geometrically complex and textured objects, the baseline methods (LAPIG\cite{deng2025lapig} and NepMap\cite{erel2023neural}) suffer from severe spatial-semantic misalignment. In addition, because they lack explicit physical guidance, their generated content frequently exceeds the reproducible color gamut of the projection surfaces, forcing the projector system to perform extensive out-of-gamut clipping. This results in darkened appearances, lost textures, and incorrect hues. In contrast, ConPhyG directly integrates depth, edge, and per-pixel gamut constraints into the generative process. We find that LAPIG is primarily constrained to near-planar surfaces and whole-image-space stylization; therefore, it exhibits limited adaptability when applied to the spatial augmentation of 3D objects. As shown in our results, the projected contents not only perfectly adhere to the structural boundaries of the 3D objects but also exhibit vibrant, accurate colors that are highly faithful to the text prompts, effectively utilizing the expanded joint gamut of the Procams setup.

\textbf{Quantitative Comparison.} It is crucial to note that the baseline methods we compared against, LAPIG and Nepmap, are inherently
constrained to single-projector configurations by their algorithmic
design. Therefore, here the setups of the three methods used the same positioned single Procams to ensure a fair evaluation of their performance. To objectively measure the projection quality, we evaluate the captured results across two dimensions: semantic fidelity to the text prompts and the radiometric color accuracy of the real projection system. As reported in Table \ref{tab:quantitative_results}, our method significantly outperforms the baselines across all metrics. For semantic alignment, ConPhyG achieves the highest CLIP score~\cite{radford2021learning} (26.37) and an advantage in BLIP ITM~\cite{li2022blip} (78.9\% compared to $\sim$17\% for baselines), indicating that our physically-guided content matches the intended language descriptions. In terms of pixel-level radiometric accuracy, our method achieves the highest PSNR of 22.38 and significantly reduces the color error ($\Delta E$)~\cite{sharma2005ciede2000} between the target images and captured images to 4.69. This reduction in $\Delta E$ confirms that our IGDN successfully prevents severe color distortion during physical display. Notably, with the same level of the maximum brightness, our method has much less color clipping error than that of the baseline methods.

\begin{table}[ht]
\centering
\caption{Comparison of projection preparation time and generation time for three methods. All experiments assume a one-second acquisition time per pattern image.}
\label{tab:efficiency}
\begin{tabular}{ccccc}
\toprule
\multirow{4}{*}{\textbf{Method}} & \multicolumn{3}{c}{\textbf{Setup}}  & \textbf{Generation}         \\ \cmidrule(lr){2-4} \cmidrule(l){5-5}
                        & \multicolumn{1}{c}{\begin{tabular}[c]{@{}c@{}}required\\ patterns\end{tabular}} & \multicolumn{1}{c}{\begin{tabular}[c]{@{}c@{}}training\\ time(s)\end{tabular}} & \begin{tabular}[c]{@{}c@{}}total \\ time(s)\end{tabular} & \begin{tabular}[c]{@{}c@{}}time(s)\\  per prompt  \qquad \end{tabular} \\ \midrule
LAPIG~\cite{deng2025lapig}                & \multicolumn{1}{c}{449}                                                          & \multicolumn{1}{c}{73}                                                         & 522                                                       & $\mathbf{7}$                                                           \\ 
NepMap~\cite{erel2023neural}                & \multicolumn{1}{c}{306}                                                          & \multicolumn{1}{c}{3500}                                                         & 3806                                                       & 15                                                           \\ 
Ours                    & \multicolumn{1}{c}{$\mathbf{45}$}                                                          & \multicolumn{1}{c}{$\mathbf{0}$}                                                         & $\mathbf{45}$                                                       & 14                                                           \\ 
\bottomrule
\end{tabular}
\end{table}

\textbf{Setup and Generation Efficiency.} Beyond display quality, practicality in real-world interactive projection mapping is a critical metric. As detailed in Table \ref{tab:efficiency}, conventional generative PM methods are often bottlenecked by tedious calibration or per-scene optimization. LAPIG and NepMap require capturing up to 449 and 306 structured light patterns, respectively, with NepMap demanding an additional 3500 seconds for NeRF/Gaussian Splatting fine-tuning (training). In contrast, our framework requires only 45 structure patterns to extract all necessary geometric correspondence and the color-mixing matrix. Specifically, we project 40 Gray code images (ten standard and ten inverted patterns for both horizontal and vertical directions) to guarantee high-precision spatial decoding in the presence of indirect reflections, alongside five uniform illumination patterns (full red, green, blue, white, and black) for accurate gamut acquisition. Furthermore, since our IGDN and diffusion model are scene-independent, they are pre-trained and entirely eliminate the need for per-scene fine-tuning. This enables ConPhyG to adapt rapidly to different projection setups, achieving a speedup of one to two orders of magnitude over previous state-of-the-art methods in terms of setup adaptation time. In the future, we plan to apply feedforward-based 3D reconstruction neural networks such as NSL~\cite{li2025robust} and encoded color mixing acquisition~\cite{Li:13} to further reduce the number of required pattern projections, potentially shortening the entire pre-processing pipeline to just a few seconds.  Regarding the per-prompt generation time, since SDXL utilizes a UNet backbone with three times the parameter count of earlier Stable Diffusion models and operates on larger $128 \times 128$ latent representations for native high-resolution synthesis, processing these expanded feature maps during denoising incurs substantial computational overhead. Therefore, it does not offer a significant advantage in generation efficiency at this stage.


\begin{figure}[t]
  \centering
  \includegraphics[width=\linewidth]{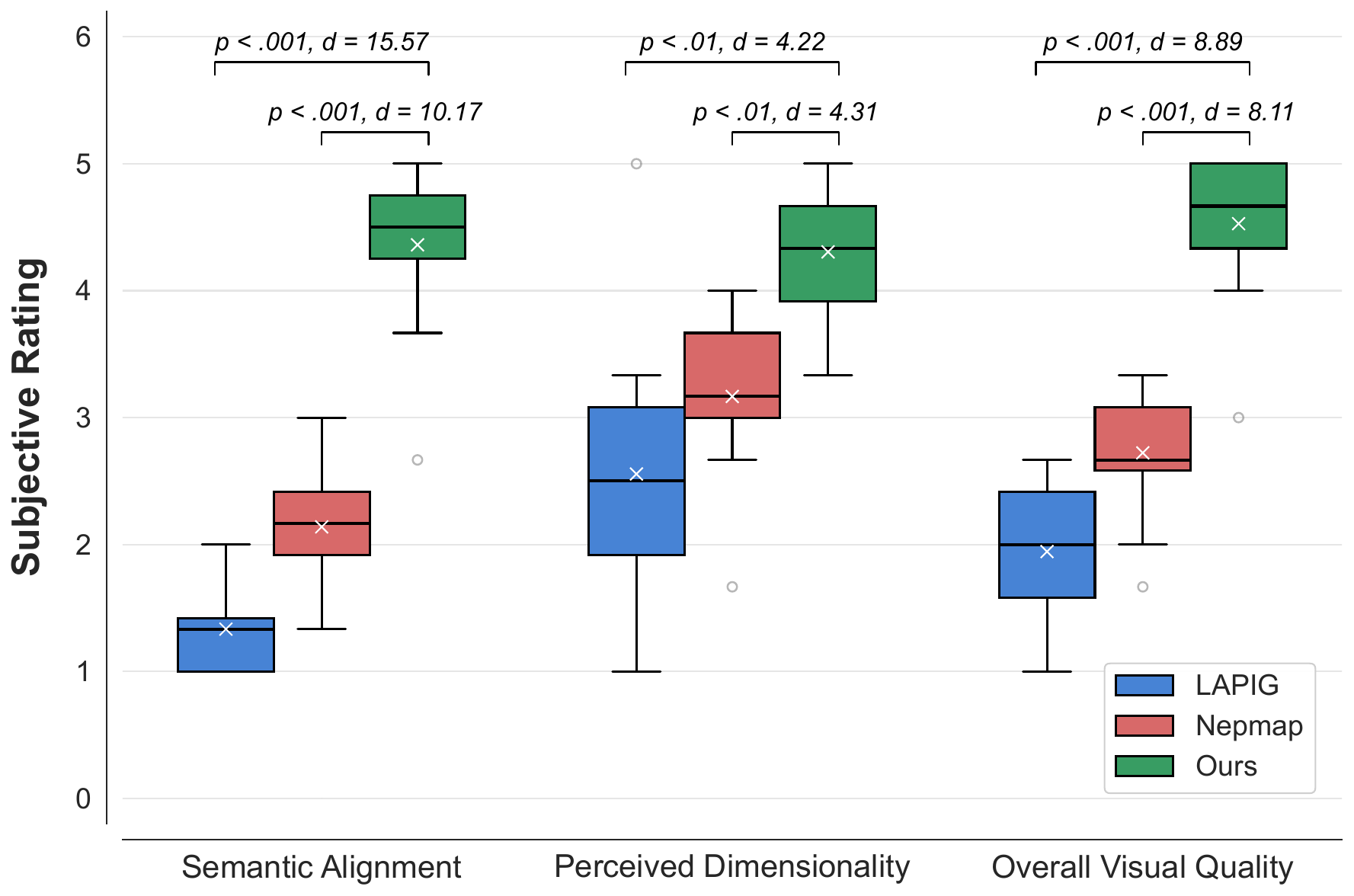} 
  \caption{Boxplots representing the user study ratings across three dimensions: Semantic Alignment, Perceived Dimensionality, and Overall Visual Quality. The center horizontal line denotes the median, while the '$\times$' symbol represents the mean. Whiskers extend to the minimum and maximum scores (excluding outliers, plotted as circles). Statistical significance from post-hoc tests is indicated by brackets above the boxes, along with the corresponding Cohen's $d$ effect sizes, highlighting the substantial perceptual advantage of our proposed method.}
  \label{fig:user_study_boxplot}
\end{figure}

\subsubsection{User Study}

\textbf{Participants and Procedure.} 
To subjectively evaluate the visual performance of our method and previous methods, we conducted an image-based perceptual study. We recruited twelve participants with normal or corrected-to-normal vision and no color blindness. 

While in-person evaluations offer physical presence, they inherently introduce confounding variables such as inconsistent viewing angles, varying ambient lighting, and observer height differences~\cite{kruijff2010perceptual}. To ensure a strictly controlled and reproducible evaluation environment, participants were presented with high-fidelity photographs of the physical projection setups, captured from the optimal viewpoint under identical lighting conditions. The baselines and our results were shown in a randomized order to prevent bias. 

For each projection result alongside its driving text prompt, participants evaluated the augmentation on a five-point rating scale across three key dimensions: 
1) \textit{Semantic Alignment} (1 = Completely Inaccurate, 5 = Highly Accurate): how accurately the visual appearance reflects the text prompt; 
2) \textit{Perceived Dimensionality} (1 = Completely Flat, 5 = Highly Dimensional): the degree to which the projected texture preserves 3D structural cues; 
3) \textit{Overall Visual Quality} (1 = Very Poor, 5 = Excellent): subjective satisfaction regarding realism, sharpness, and the absence of visual artifacts~\cite{wang2004image}.

\textbf{Results and Statistical Analysis.}
We analyzed the collected rating data to investigate the effect of different projection mapping methods on user perception. Assuming the 5-point rating scale data approximates interval variables, we conducted a one-way repeated measures ANOVA for each of the three evaluation dimensions. Post-hoc pairwise comparisons were performed using paired t-tests with Bonferroni corrections to control for Type I errors. The statistical results are visualized in Fig.~\ref{fig:user_study_boxplot}.

\textit{Semantic Alignment:} The RM-ANOVA revealed a statistically significant main effect of the projection method on Semantic Alignment ($F(2, 22) = 124.76, p < .001$). Post-hoc tests demonstrated that our proposed method ($M = 4.36, SD = 0.67$) scored significantly higher than LAPIG ($M = 1.33, SD = 0.32, p < .001, d = 15.57$) and NepMap ($M = 2.14, SD = 0.52, p < .001, d = 10.17$). This confirms that explicitly incorporating physical gamut constraints allows our method to better preserve the semantic color intent of the text prompts, overwhelmingly outperforming existing baselines.

\textit{Perceived Dimensionality:} A significant main effect was also found for Perceived Dimensionality ($F(2, 22) = 14.84, p < .001$). Participants rated the 3D structural fidelity of our method ($M = 4.31, SD = 0.54$) significantly better compared to both LAPIG ($M = 2.56, SD = 1.11, p = .001, d = 4.22$) and NepMap ($M = 3.17, SD = 0.61, p = .001, d = 4.31$). This indicates that the inclusion of depth and edge-aware physical priors successfully prevents the virtual content from appearing as a flattened texture, effectively preserving the object's original geometric cues.

\textit{Overall Visual Quality:} For the overall visual satisfaction, the RM-ANOVA similarly indicated a highly significant main effect ($F(2, 22) = 63.28, p < .001$). Post-hoc comparisons confirmed that our approach ($M = 4.53, SD = 0.58$) was significantly preferred over LAPIG ($M = 1.94, SD = 0.62, p < .001, d = 8.89$) and NepMap ($M = 2.72, SD = 0.53, p < .001, d = 8.11$). Users frequently reported that our projections appeared sharper and suffered from fewer clipping artifacts, validating the practical efficacy of our end-to-end physically-guided generative pipeline.


\begin{figure*}[htp]
  \centering
  \includegraphics[width=\linewidth]{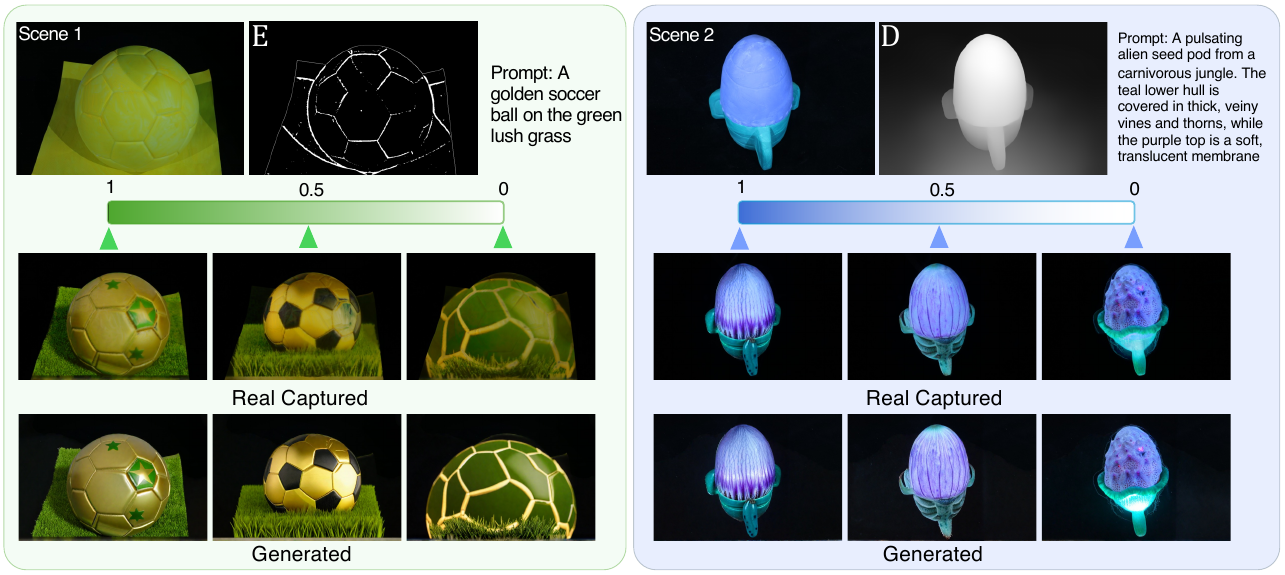}
  \caption{Ablation Study on geometric constraints. We progressively reduce the conditioning strength of the edge map (Left) and depth map (Right) to evaluate their individual contributions to spatial augmentation. \textbf{(Left) Edge Map Ablation:} As the edge strength decreases from $1.0$ to $0$, the generated texture progressively loses its precise alignment with the physical boundaries (e.g., the generated content fails to match the seam of the real soccer ball). \textbf{(Right) Depth Map Ablation:} Reducing the depth conditioning strength causes the generated 3D structures to deviate significantly from the underlying geometry. While the upper capsule shape is partially preserved by residual edge guidance, the intricate vine structures at the base undergo severe structure distortion. Notably, at a depth strength of $0$, the generated content exhibits drastic volumetric shrinkage, leaving ghosting artifacts (faint outlines) at the periphery.}
  \label{fig:ablation}
\end{figure*}

\begin{figure}[htp]
  \centering
  \includegraphics[width=\linewidth]{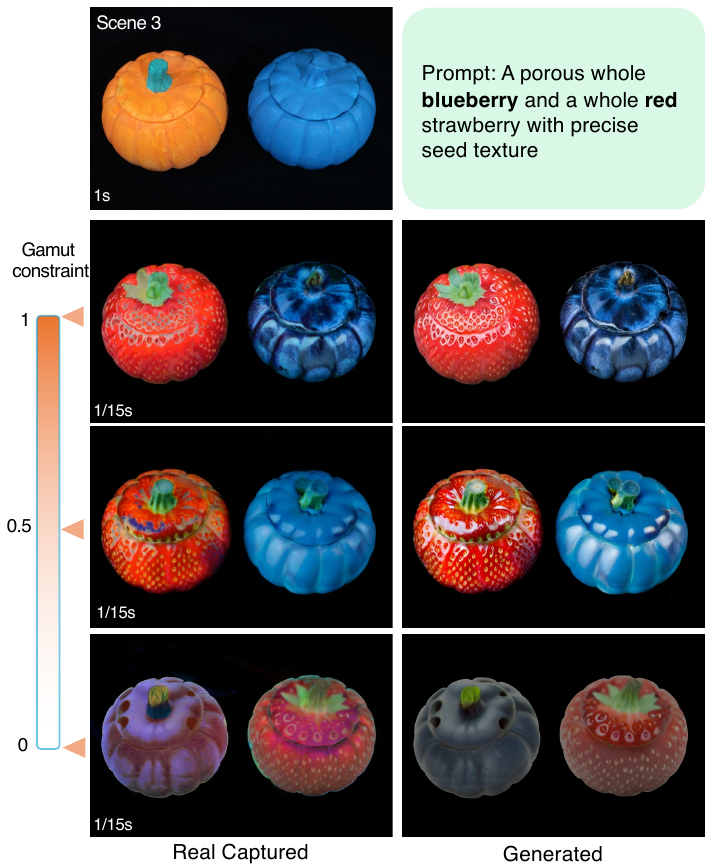}
  \caption{Ablation of gamut constraint under varying strength. Progressively decreasing the gamut constraint strength from 1.0 to 0.0 relaxes the physical boundaries, causing generated colors to increasingly violate the object's surface reflectance properties. This degradation results in severe color clipping and brightness compression, such as generating red strawberries on blue surfaces. }
  \label{fig:gamut ablation}
\end{figure}

\begin{figure}[tb]
  \centering
  \includegraphics[width=\columnwidth]{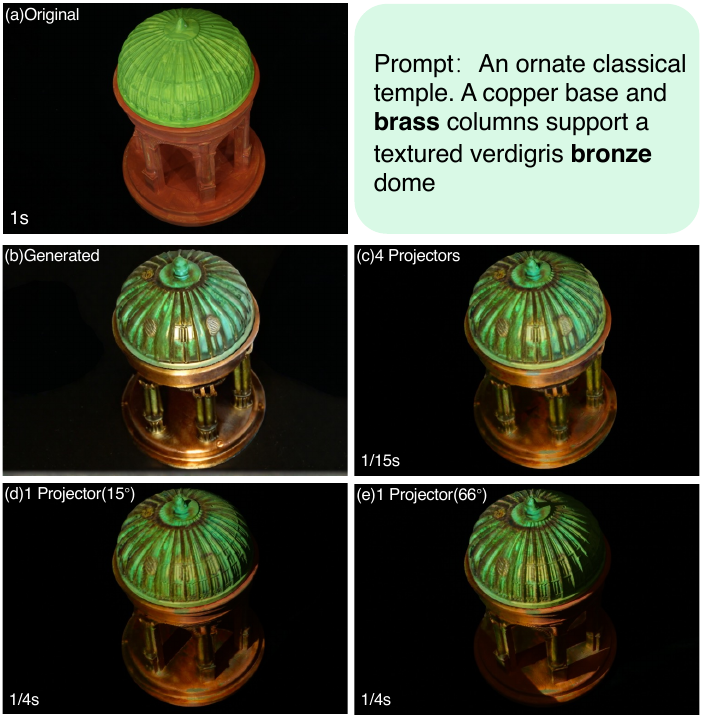}
  \caption{Qualitative comparison of spatial augmentation across different projector configurations. (a) The original, unaugmented physical object. (b) The ideal generated target appearance. (c) Our four-projector setup delivers uniform luminance and comprehensive spatial coverage, accurately reproducing the target texture on the object. (d) and (e) Single-projector setups restrict the augmentation to a unilateral surface. Furthermore, as the deployment angle of the single projector increases relative to the camera's optical axis (from 15° to 66°), the projection quality degrades sharply. These single-projector configurations suffer from severe self-occlusion and hard shadows caused by the object's own geometric complexity.}
  \label{fig:4ProsVS1Pro}
\end{figure}

\subsection{Ablation Study}

To evaluate the contribution of each core component in the ConPhyG framework, we conduct systematic ablation experiments focusing on geometric constraints (depth and edge) and the radiometric constraint (physical gamut guidance).

\textbf{Quantitative results} To quantify the contribution of each physical constraint, we evaluate perceptual depth, geometric alignment, and color fidelity. We calculate the normalized Root Mean Square Error (RMSE) of the depth map and the Chamfer Distance between the structural edges map, which evaluates the alignment accuracy across 3D topologies and 2D contours. For the gamut constraint, we compute the CIE 2000 color difference ($\Delta E$) between the generated images and the actual camera-captured images to assess color fidelity and out-of-gamut truncation error.

As shown in Table~\ref{tab:ablation_constraints}, omitting any physical prior significantly degrades projection quality. Without gamut guidance (w/o gamut), generated colors deviate from the surface's reproducible gamut, forcing extensive color clipping that increases $\Delta E$ from 4.69 to 7.01, aligning with our observed qualitative distortions. Geometrically, removing depth conditioning (w/o depth) strips the virtual textures of their underlying 3D surfaces geometry, worsening RMSE from 0.09 to 0.36. Similarly, disabling the edge constraint (w/o edge) causes severe boundary misalignment, drastically increasing the Chamfer Distance from 105 to 535. These results confirm that our proposed physical modules are essential for bridging the generative and physical projection domains.

\begin{table}[htbp]
    \centering
    \caption{Quantitative Evaluation (color: $\Delta E$, depth: \textbf{RMSE}, edges: \textbf{Chamfer Distance}) of Ablation Study.} \label{tab:ablation_constraints}
    \resizebox{\columnwidth}{!}{%
    \begin{tabular}{lccc}
        \toprule
        \textbf{Method} & $\Delta E \downarrow$ & \textbf{RMSE} $\downarrow$ & \begin{tabular}[c]{@{}c@{}}\textbf{Chamfer}\\ \textbf{Distance}\end{tabular} $\downarrow$ \\
        \midrule
        w/ gamut, depth and edge  & \textbf{4.69} & \textbf{0.09} & \textbf{105} \\
        w/o gamut                 & 7.01  & 0.11 & 121 \\
        w/o depth                 & 4.77  & 0.29 & 301 \\
        w/o edge                  & 5.69  & 0.16 & 422 \\
        w/o depth and edge        & 6.03  & 0.35 & 628 \\
        w/o gamut, depth and edge & 35.33 & 0.44 & 1907 \\
        \bottomrule
    \end{tabular}%
    }
\end{table}

\textbf{Effect of Geometric Constraints.}
We analyze the individual impacts of the texture edge map and depth map on the spatial alignment and structural fidelity of projected content.
For the edge constraint, we progressively reduce the conditioning strength of the Canny edge map from 1.0 to 0. Fig. \ref{fig:ablation} provides a representative example. Full edge conditioning ensures precise preservation of physical object boundaries, while reduced strength leads to severe texture misalignment and edge bleeding in projected results.
For the depth constraint, we similarly adjust the conditioning strength of the depth map. Fig. \ref{fig:ablation} also provides a representative example. Full-depth conditioning enables the generation of an accurate 3D geometry appearance that conforms to the target surface shape. Without depth conditioning, the model produces flattened or structurally incorrect geometries that ignore the physical curvature of the object, severely degrading the 3D perceptual illusion.

\textbf{Effect of Physical Gamut Guidance.}
To isolate the effectiveness of the IGDN-based gamut guidance module, we perform a controlled experiment where generation is initialized with a uniform neutral-gray canvas (R=G=B=128) at full denoising strength (1.0). This setup eliminates all latent color priors from input images, ensuring semantic color routing is driven solely by our gamut backpropagation mechanism.
Without the gamut constraint, the model generates textures based exclusively on text prompts, disregarding the surface's spectral reflectance characteristics. This forces the intensity solver to perform extensive color clipping to meet physical constraints, resulting in darkened and distorted projected results. As shown in Fig. \ref{fig:gamut ablation}, when attempting to generate a strawberry on a blue pumpkin and a blackberry on an orange pumpkin, the model without gamut guidance produces severely desaturated outputs. Conversely, with gamut guidance enabled, the IGDN continuously backpropagates physical feasibility gradients to steer latent semantics toward the projectable color range. This ensures generated content remains within the Procams setup's reproducible gamut, producing vibrant and semantically faithful results.

\textbf{Effect of Multi-Projector Configuration.}
To verify the necessity of a multi-projector setup for single-viewpoint spatial augmentation, we compare our default four-projector configuration against single-projector baselines deployed at different angles.
Although both setups target the same primary viewing direction, the four-projector configuration provides uniform luminance and comprehensive spatial coverage of the visible 3D surface, effectively eliminating geometric blind spots, as shown in Fig. \ref{fig:4ProsVS1Pro}. In contrast, single-projector setups fail to fully augment complex surfaces, with projection quality degrading sharply as the deployment angle relative to the camera's optical axis increases. Single-projector configurations suffer from severe self-occlusion and prominent hard shadows caused by the object's geometric intricacies, which disrupt visual continuity and break the unified augmented texture illusion.


\begin{figure*}[htp]
  \centering
  \includegraphics[width=\linewidth]{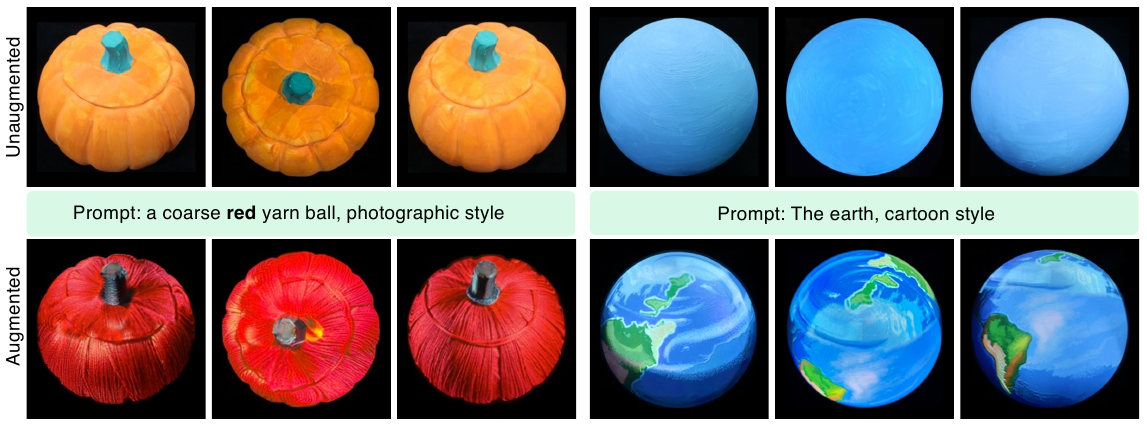}
  \caption{Visualization results of our preliminary 360-degree view-independent projection mapping. (Upper: objects and prompts. Bottom: captured images) By employing the sequential compensation strategy and grazing-angle mask erosion, our framework achieves semantic consistency and geometric alignment across multiple viewpoints(two side views and one top view), effectively bridging the gap between digital generative content and 3D physical surfaces.}
  \label{fig:360_results}
\end{figure*}

\subsection{Evaluation of 360-Degree Spatial Augmentation}
To demonstrate the potential of our framework, we apply our 360-degree view-independent projection mapping strategy to axisymmetric objects and present the visualization results in Fig.~\ref{fig:360_results}. As observed in the multi-view captured images, our sequential generation strategy successfully augments the real objects according to the semantic themes defined by the text prompts. Even when the observer’s viewpoint shifts significantly, the projected textures remain structurally coherent. Although 2D content generation across disparate viewpoints inherently suffers from cross-view inconsistencies, the target objects used in this study are relatively simple, characterized by axisymmetric geometries and uniform surface reflectance. By leveraging previously synthesized views as conditional guidance within a sequential generation pipeline, our framework effectively mitigates these inconsistencies, achieving seamless 360-degree projection mapping results on such objects.

We also observe that the omnidirectional illumination inherent in the 360-degree setup inevitably introduces global illumination effects. These inter-reflections introduce subtle radiometric artifacts and contrast degradation that are less pronounced in the single-view configuration.
\section{Discussions and Limitations}

\subsection{Discussions}

Our method is a unified framework that flexibly switches generation strategies according to constraint conditions to accommodate both paradigms. When physical constraints are enabled, our framework follows the cooperative principle by embedding object priors into the diffusion model to generate surface-conformal projection content. When no object constraints are imposed, the framework first produces unconstrained creative content and then conducts multi-projection radiometric compensation. Nevertheless, currently our method still lacks the ability to adaptively select the optimal operating point across all alignment dimensions and cannot automatically switch between paradigms according to real-time alignment conditions between user intent and physical objects. We believe that realizing such an adaptive mechanism will bring a paradigm shift to projection mapping research, enabling the technique to adapt to diverse user creative intentions rather than limiting creative freedom. This future improvement can further provide a new theoretical foundation for developing more flexible and efficient spatial augmented reality systems.

\textbf{Acceleration.} 
Improving the computational efficiency of the entire projection mapping pipeline is crucial for its practical deployment, especially in interactive application scenarios. For content generation acceleration, state-of-the-art models such as SANA~\cite{xie2024sana} have shown promising high-resolution results with generation times in a second, which can potentially reduce content generation time for our framework in future work.
Another key acceleration direction lies in reducing the acquisition of object geometry and reflectance. By adopting more compact structured light techniques~\cite{zhang2018high}, redundant encoding in the acquisition process can be effectively reduced, enabling the system to obtain the required geometric data and color mixing matrix in a shorter time while maintaining sufficient measurement accuracy.

\subsection{Limitations}

Our method still faces the following three major technical limitations:

\textbf{2.5D Multi-Projection Mapping.} While our approach accounts for surface depth and enables view-independent PM, it lacks explicit multi-view consistency constraints. We therefore cautiously classify our method as 2.5D multi-projection mapping.
Notably, existing generative methods also fail to produce valid multi-view PM, as they also suffer from inconsistencies in the overlapping regions of different views. This is an inherent limitation of applying 2D image generation techniques to 3D object surfaces.
This issue could potentially be addressed by incorporating 3D generative models in future work. At present, however, such integration remains impractical due to prohibitively high computational costs and relatively low generation quality. For future work, we will explore different 3D content generation techniques, such as Score Distillation Sampling (SDS)~\cite{poole2022dreamfusion}, and seamlessly embed the generation model into our framework to enhance multi-view consistency while ensuring both computational efficiency and projection quality.

\textbf{Channel-Independent Assumption.} 
In the experiments, we employ multiple channel-independent LCD projectors and pre-calibrate their non-linear response function. This allows the multi-primary color gamut model to be drastically simplified, making color sampling easy to implement. As reported in previous work~\cite{sajadi2010adict}, the color gamuts rendered by DLP projectors on the surfaces with saturated color reflectance exhibit strong concavity. The more complex projector response functions would introduce additional computational burden for in-gamut determination. Fortunately, previous studies have demonstrated that either the nonlinear Thin Plate Spline–radial basis function (TPS-RBF)~\cite{grundhofer2013practical} or a four-layer MLP~\cite{li2023physics} can effectively represent the projector response. Such a nonlinear representation can thus also be adopted to model the projector color gamut. Future work will focus on relaxing the channel-independent assumption, extending the MLP-based representation to model coupled channel responses, and validating the improved model with channel-coupled projectors.

\textbf{Direct Illumination.} 
A notable limitation of our current framework lies in the radiometric compensation pipeline. After generating the desired camera image, we directly adopt the pre-acquired per-pixel color mixing matrix paired with our in-gamut determination network to solve for the per-pixel intensity of the projector image. This choice is driven by maintaining high computational efficiency and preserving the reusability of our pre-trained in-gamut determination network. However, this simplified pipeline does not explicitly model the indirect illumination effects inherent to real-world projector-camera systems, including projector defocus blur, subsurface scattering, and inter-reflection between scene surfaces. These effects may lead to suboptimal radiometric compensation performance in complex projection scenarios. For future optimization, projector defocus blur can be effectively mitigated using the extended depth of field (EDOF) method with learned optics~\cite{li2023extended}, while the adverse effects of subsurface scattering and inter-reflection can potentially be compensated via a Transformer network with attention mechanism~\cite{zeng2025renderformer}.

\section{Conclusion}
This paper introduces ConPhyG, the first controllable text-driven, physically-guided framework for multi-projection mapping content generation. By embedding physical feasibility constraints of projectors and target objects (e.g., per-pixel gamut, depth, texture edges) directly into the text-driven generative pipeline, paired with our novel in-gamut determination network for gamut-aware content creation, our framework guarantees consistent, harmonious fusion between real and virtual content. Experiments and user studies verify that our method achieves robust compatibility with multi-projector overlap setups and textured target surfaces, while consistently outperforming state-of-the-art methods in setup efficiency, display brightness, color fidelity, and operational flexibility. This work establishes a unified, low-barrier paradigm for accessible, high-quality projection mapping content creation. In future work, we will explore true 3D projection mapping and enable adaptive constraints within our framework to further expand its technical capabilities.

\bibliographystyle{IEEEtran}
\bibliography{template}

\vfill

\end{document}